\documentclass[submission,copyright,creativecommons]{eptcs}
\providecommand{\event}{AREA 2023} 

\usepackage{iftex}

\ifpdf
  \usepackage{underscore}         
  \usepackage[T1]{fontenc}        
\else
  \usepackage{breakurl}           
\fi

\usepackage{graphicx}
\usepackage{float}

\usepackage[dvipsnames]{xcolor}

\usepackage{academicons}
\definecolor{orcidlogocol}{HTML}{A6CE39}

\title{Autonomous Systems' Safety Cases for use in UK Nuclear Environments\thanks{This work is supported by the EPSRC, through the Robotics and AI for Nuclear (RAIN) Hub (EP/R026084, EP/W001128/1, EP/R026084/1). Thanks are due to the Office for Nuclear Regulation (ONR) and Sellafield Ltd. for input into and comments upon the Safety Case this paper describes.}}

\author{Christopher R. Anderson 
\institute{Department of Computer Science\\
University of Manchester\\Manchester, UK}
\email{chris.anderson@manchester.ac.uk}
\and
Louise A. Dennis 
\institute{Department of Computer Science\\
University of Manchester\\Manchester, UK}
\email{\quad louise.dennis@manchester.ac.uk}
}

\def\titlerunning{Autonomous Systems' Safety Cases for use in UK Nuclear Environments}
\def\authorrunning{C.R. Anderson, L.A. Dennis}
\begin{document}
\maketitle              

\begin{abstract}
An overview of the process to develop a safety case for an autonomous robot deployment on a nuclear site in the UK is described and a safety case for a hypothetical robot incorporating AI is presented. This forms a first step towards a deployment, showing what is possible now and what may be possible with development of tools. It forms the basis for further discussion between nuclear site licensees, the Office for Nuclear Regulation (ONR), industry and academia.

\end{abstract}

\section{Introduction}

Autonomy lends itself to activities in the nuclear industry. Traditional, verifiable~\cite{61508,60880} robotic systems lack the ability to perform many desirable tasks, however, this could be made possible by the use of Artificial Intelligence (AI) technologies, which include machine learning (ML) (sub-symbolic) and formally verifiable logical reasoning (symbolic). 

The deployment of these systems in nuclear environments necessitates that a complete and coherent set of arguments is made which demonstrates that the activity to be undertaken (by the autonomous system) is adequately safe, by utilising a \textbf{claims}, \textbf{arguments} and \textbf{evidence} trail~\cite{ONR2014} (CAE).

However, in the view of the UK nuclear industry, AI is not a mature technology that can easily be shown to meet the well-established, conservative approaches to safety. It can appear difficult or indeed impossible to construct a safety case, in contrast to other domains where AI is in the process of being adopted (e.g. automotive).  Whilst the nuclear consequence may be higher, the environment in which the activity is undertaken is generally well constrained. We have proposed a route that, with some thought and good engineering practices, can enable a safety case to be constructed.  This builds on a white paper providing principles for the development and assurance of autonomous systems~\cite{Luckcuck2021}. An outline architecture and safety case~\cite{Anderson2022} has been developed to demonstrate this.

\section{The Robot and Task}
In the absence of a suitable and timely deployment, a hypothetical robot and scenario have been used based very loosely on the A2I2 Lilypad ASV \cite{Watson2021} deployed to survey a nuclear waste storage pond to assist with remediation tasks. These ponds hold the spent fuel from a reactor in water which has a depth of more than 10 m. The robot is programming using a combination of symbolic and sub-symbolic AI.

A Robotics with Autonomy/AI Safety Case framework for our hypothetical robot can be found on-line~\cite{Anderson2022}\footnote{Note that, this Safety Case was generated using an ASCE 
Academic Licence and therefore cannot be used for commercial purposes}.  In this paper we discuss salient features of this safety case that would enable similar safety cases to be created, for real deployments in UK Nuclear environments.

\section{Safety Cases in the UK Nuclear Industry}
In this section we discuss the key features of creating a safety case for use in a UK Nuclear environment, with particular focus on those aspects relevant to the deployment of autonomous robotic systems.

\textbf{Identification of hazards:}
The building of a safety case starts with the holistic identification of hazards, analysed for the robot within its application task and environment, whether the robot is pre-existing or proposed. 
Analyses of robot behaviour that do not account for the specific task and location are not sufficient on their own for creating a safety case.
Hazards should be identified for all of the robot's lifecycle phases and tasks.  This can usually be achieved by applying the site licensee’s high level principles. Normal and abnormal operations should also be analysed.

\textbf{Developing risk mitigation strategies:}
Mitigation strategies are  derived through one or more optioneering studies (the selection of the best option from a set of alternatives), where the objective is to reduce risks to As Low As Reasonably Practicable (ALARP). This process should identify the benefits and disadvantages of any proposed solution in comparison to other options.  It may include an argument that deployment of a prototype technology that may have not be preferable for the specific task has  long term benefits that out-weigh the immediate disadvantages.  The aim is to either reduce the consequence (which is typically fixed) or reduce the likelihood of any unwanted outcome. 
Defence in depth principles are applied to each hazard.  In the case of autonomous robots, most of the robotics design efforts lie in an ‘occurrence barrier’ regime -- the protection and mitigation barriers are usually external to the robot. 

This paper assumes that an optioneering study has determined that an autonomous robot is necessary. 

\textbf{Tolerability and ALARP:}
The principles of tolerability and ALARP are key to acceptance of the autonomous robot to perform a particular task at a particular location on a nuclear site. Tolerability is expressed in The Tolerability of Risks from Nuclear Power Stations (TOR), 1992 \cite{Ballard1992} which defines risks which are so high they are unacceptable unless there are exceptional circumstances. The requirement for risks to be ALARP (take all measures to reduce risk where doing so is reasonable \cite{ONR2020}) is fundamental and applies to all activities within the scope of the Health and Safety at Work Act, 1974 (HSWA). ALARP can be achieved through applying established ‘Relevant Good Practice’ (RGP) and standards and only in cases where these are inappropriate is a cost/benefit analysis used. Tolerability and ALARP are understood through the diagrams in TAG 094~\cite{ONR2019}, Figure 3.

\textbf{Safety functions}
A Safety Function (SF) is a mechanism for implementing mitigation strategies.  An SF  can be realised as either: a function which is diverse, independent and segregated from the control system, sensors, control and actuators of the robot (a guard); the control function (system) itself or a combination of guard and control system. The guard and/or control system must: lend itself to design, implementation, verification \& validation to the degree required by the hazard analysis and the safety requirements (functional and non-functional) imposed on it; meet all non-probabilistic requirements; meet the probabilistic claim required by the hazard analysis and the safety requirements imposed on it.

Safety Functions (SF) are realised by Structures, Systems and Components (also known as Safety Instrumented Functions (SIF)), using appropriate standards and RGP. e.g. IEC 61508 \cite{61508}. Production Excellence (PE), one of two “legs” used to substantiate safety to the regulator, demonstrates good control of the robot’s development and verification lifecycle. The second leg, Independent Confidence Building Measures (ICBM), requires that the final validated software (in its target hardware deployment) and the testing programme be independently checked and can include statistical testing and static code analysis. In general, it is always preferrable to use the simplest, most effective mitigation and this should be demonstrated in the optioneering study and ultimately in the safety case.

\section{The Safety Case}
It is widely recognised that the use of autonomy/AI in robotics in high integrity applications can be difficult to justify.  In particular challenges in analysing the behaviour of such systems form a challenge to the creation of a robust safety case. The definition of the activity and formal identification and analysis of the hazards (the Preliminary Hazard Analysis (PHA)) forms the primary, and probably most crucial, task in such a robot’s development lifecycle. The PHA ensures that the safety system is not over engineered or, indeed, unnecessary. In particular, we assert that it is not always necessary to introduce new mitigations simply because autonomy is involved.

Whilst there are potentially a number of hazards which the robot we consider here could either encounter or initiate, including propeller splash, being irretrievable due to complete robot system failure and explosion due to hydrogen evolution at the surface of the pond, the safety case presented here focuses on collision with either the pond structure or its contents and we assume that the PHA has resulted in the identification of this as a hazard that requires mitigation.

Our safety case has been generated as an example framework, based on a hypothetical robot operating in a nuclear material storage pond.  Consequently, in several places the safety case contains placeholder nodes since this documentation is not available for the hypothetical robot. The safety case provides two examples of how a `Collision' hazard may be mitigated (referred to below as Method 1 and Method 2). It is unlikely that both methods would be needed to show that the risk has been reduced to be ALARP and, therefore, one could be deleted. In the event that both are needed to show defence in depth and diversity, then Method 1 and 2 would need to be restructured into one set of CAE. We are reasonably certain that Method 1 can be used now, and it is feasible that Method 2 can be used in the medium term, providing the necessary verification tools and methodologies are developed.

The ‘Top View’ provides the best entry point for the safety case~\cite{Anderson2022}. Here the primary claim (node C1) “Robot is adequately safe... to provide...” is elaborated by the main elements: Optioneering study and justification of the choices made; establishment and analysis of the hazards; the derivation of safety requirements; compliance of the design with safety requirements; the use within a safety management system; and compliance with standards and RGP. The hazards are separated into four groups: Nuclear (radiological), 
conventional, 
physical, 
and cyber security. 
The compliance of the design with safety requirements is further broken down and eventually reaches the claim that the safety risk is mitigated and managed. 
At this claim node the arguments split into Method 1 (Using an Engineered SSC in the form of a guard) 
and Method 2 (Using verifiable AI technologies). 

\textbf{Robot hazard and mitigations:}
The collision hazard for our hypothetical robot was determined to have a consequence of $<$ 2 mSv (Sievert - the SI (International System of Units) unit which represents the stochastic risk to health of ionising radiation), based on experience of this environment. However, normally specific site licensee experience would be applied here. The principle of ALARP means that the same strategies can be applied in the range of 2 to 20 mSv (below the BSL for on-site workers). We deem this hazard to be mitigated by the use of an Occurrence Barrier which we describe below.

\subsection{Occurrence barrier (SIF)}
This safety case proposes two approaches to mitigation of the collision hazard, as follows:

\textbf{Diverse guard (\cite{Anderson2022}, Method 1:} 
This SIF is comprised of cat’s whiskers surrounding the robot which actuate microswitches and in turn, open safety relay contacts, removing the delivery of power to the propellers. This is relatively crude and does not allow for subtle control. e.g. an ‘intelligent’ robot may be able to avoid the collision by steering away from the obstacle or reversing away from it. From a safety perspective it has the advantage of the ability to make a highly deterministic and probabilistic claim because of its simplicity. A minimal claim is necessary to ensure that the SIF (guard) is not demanded too often thereby causing excessive deterioration of the components or encouraging the operator to ignore or disable the safety feature.

\begin{figure}
    \centering
    \includegraphics[width=0.91\textwidth]{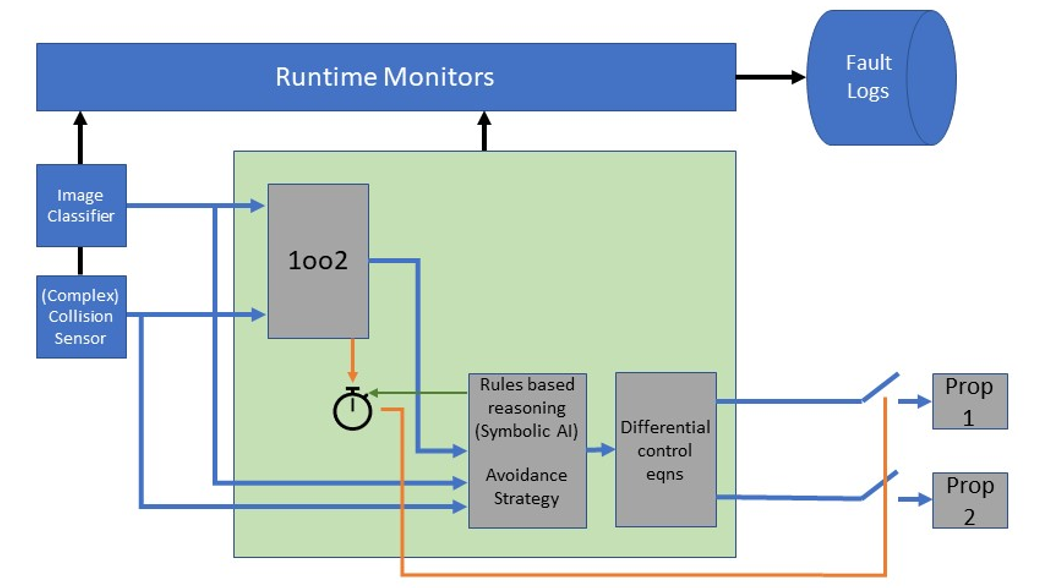}
    \caption{Rules based reasoning SIF (Method 2).} \label{Rules based reasoning SIF (Method 2)}
    \label{fig:method2}
\end{figure}

\textbf{Rules based reasoning (\cite{Anderson2022}, Method 2:} 
This SIF is comprised of the intelligent (safety) control system itself and a collision sensor with a software component, as shown in Figure \ref{Rules based reasoning SIF (Method 2)}. At higher Safety Integrity Level (SIL) (lower Class in nuclear) this may be completely independent of any ML-based components involved in implementing the base functionality of the robot. We assume that navigation utilises an ML image classifier and that other ML components may have been involved in developing navigation planning systems.  It is not feasible to place a claim on such systems at this time, therefore, a second independent (software based) sensor, such as an ultrasonic sensor or LIDAR, is used to detect collisions on which a safety claim is made. This and the differential control equations that control motion have been constructed from traditional software and hardware and developed using a safety lifecycle process such as that described in IEC 61508 \cite{61508} at the appropriate SIL for which PE and ICBM can be demonstrated. 

The detection of the potential collision and actuation of the consequential avoidance action is then mediated by a 1-out-of-2 voting system~\cite{Storey1996}, between the image classifier and the collision sensor, providing output to a separate intelligent control system. The intelligent control system here is constructed from a rules based reasoning (RBR) architecture (such as those supported by the MCAPL framework~\cite{dennis17gwen,dennis18:mcapl}). This strategy allows for a more subtle control where the robot can take evasive action. RBR systems controlling autonomous and robotic systems work in a cyclic fashion first sensing the environment, then applying rules to pick an appropriate action and then acting before returning to sensing.  They can generally be transformed into decision tree structures but the representation as a set of rules with specified applicability conditions is more compact and so has advantages in terms of code readability.  The applicability conditions and output actions of the rules themselves define a state space which can be explored using techniques such as model-checking~\cite{Clarke00:MC} to check for desirable properties.  These properties can be expressed in a variety of temporal and probabilistic logics and tools exist to allow requirements to be captured and expressed in these logics.  Typically model-checking operates on an abstract model of the system to be verified, but so called \emph{program model-checkers} exist which apply the same process to the actual code. For instance the MCAPL Framework uses a version of the JPF model-checker~\cite{VisserM05} to verify the behaviour of RBR systems written in a number of agent programming languages.  To achieve this it executes the actual code in order to determine the next state of the system. Some property languages allow timing constraints to be expressed, such as UppAal~\cite{bllpw:dimacs95} which has a property specification language but it is not a program model-checker and we are not aware of any program model-checkers that allow timing constraints to be expressed. However, a timer and runtime monitors (and watchdog timers (WDT)) could be provided to ensure that the collision avoidance takes place in a timely fashion and triggering some default safe behaviour (e.g. the complete stop of method 1) if it did not. The full architecture is shown in Figure~\ref{fig:method2}.  To achieve this both a deterministic (complying with standards and RGP) and probabilistic (from, for example, a Failure Modes, Effects and Criticality Analysis) claim must be placed on the intelligent control system. RBR enables the use of formal verification (i.e. the application of a formal mathematical proof) of the collision avoidance. Its implementation is on an end to end basis. i.e. from requirements through to hardware deployment.  

At present the necessary technology at a high enough Technology Readiness Level (TRL) does not exist to realise Method 2.  The MCAPL Framework which enables verifiable RBR for agent programs is academic software and depends upon the underlying use of the Java programming language.  Obviously it would be preferable to use a safety critical realtime operating system and programming language.  However, the barriers to this method are primarily the availability of appropriate languages and toolsets, not a lack of methodology.

\section{Conclusion}
The safety case described here and found in \cite{Anderson2022} provides a first attempt at an argument for deploying a robot with autonomy on a nuclear site in the UK. It introduces two approaches: one of which is possible now but does not allow all of the benefits of the autonomy to be realised and a second which identifies a potential route to incorporating these benefits.  The safety case shows that autonomy in and of itself is not necessarily a barrier to the deployment of autonomous robots in UK nuclear environments and, in fact, that existing approaches to analysing hazards, devising mitigations and establishing safety claims are still applicable where autonomy is involved.  However we have also shown that there remains a gap in situations where we might want the autonomy itself to form a part of the SF and claim.  Method 2 outlines a potential approach to bridging this gap.\\

\noindent\textbf{Open and Data Access Statements:}
For the purpose of open access, the authors have applied a Creative Commons Attribution (CC BY) licence to any Author Accepted Manuscript version arising.




\bibliographystyle{eptcs}
\bibliography{references}

\end{document}